\documentclass[11pt]{article}
\usepackage{coling2018}
\usepackage{times}
\usepackage{url}
\usepackage{latexsym}

\usepackage{graphicx}
\usepackage{algorithmic}
\usepackage{algorithm}
\usepackage{amsmath, amsfonts}
\usepackage{savesym}
\usepackage{multirow}
\usepackage{booktabs}

\newcommand{\bvec}[1]{\mbox{\boldmath $#1$}}
\title{Addressee and Response Selection for Multilingual Conversation}

\author{
Motoki Sato$^1$\thanks{This work was conducted when the first author worked at Nara Institute of Science and Technology, and portions of this research were done while the third author was at IBM Research - Tokyo.}
\hspace{1cm}Hiroki Ouchi$^{2, 3}$
\hspace{1cm}Yuta Tsuboi$^{1*}$\\
$^1$Preferred Networks, Inc.,\\
$^2$RIKEN Center for Advanced Intelligence Project\\
$^3$Tohoku University\\
{sato@preferred.jp}, {hiroki.ouchi@riken.jp}, {tsuboi@preferred.jp}
}

\date{}

\begin{document}
\maketitle
\begin{abstract}
    Developing conversational systems that can converse in many languages is an interesting challenge for natural language processing.
    In this paper, we introduce {\it multilingual addressee and response selection}.
    In this task, a conversational system predicts an appropriate addressee and response for an input message in multiple languages.
    A key to developing such multilingual responding systems is how to utilize high-resource language data to compensate for low-resource language data.
    We present several knowledge transfer methods for conversational systems.
    To evaluate our methods, we create a new multilingual conversation dataset.
    Experiments on the dataset demonstrate the effectiveness of our methods.
\end{abstract}

\section{Introduction}
\label{sec:1}

\blfootnote{
    %
    %
    %
    %
    %
    %
    \hspace{-0.65cm}  
    This work is licensed under a Creative Commons 
    Attribution 4.0 International License.
    License details:
    \url{http://creativecommons.org/licenses/by/4.0/}.
}

Open-domain conversational systems, such as chatbots, are attracting a vast amount of interest and play their functional and entertainment roles in real-world applications.
Recent conversational models are often built in an end-to-end fashion using neural networks, which require a large amount of training data \cite{vinyals:2015,serban:2016}.
However, it is challenging to collect enough data to build such models for many languages.
Consequently, most work has targeted high-resource languages, such as English and Chinese \cite{shang:2015,serban:2016}.

In this work, we aim to develop {\it multilingual} conversational systems that can return appropriate responses in many languages.
Specifically, we assume the two types of systems: (i) {\bf language-specific} systems and (ii) {\bf language-invariant} systems.
A language-specific system consists of multiple conversational models, each of which returns responses in a corresponding language.
By contrast, a language-invariant system consists of a single unified model, which returns responses in all target languages.
A key to building these multilingual models is how to utilize high-resource language data to compensate for low-resource language data.
We present several knowledge-transfer methods.
To the best of our knowledge, this is the first work focusing on low-resource language enablement of conversational systems.

One challenge when developing conversational systems is how to evaluate the system performance.
For generation-based conversational systems, which generate each word for a response one by one, many studies adopt human judgments.
However, it is costly and impractical to adopt this evaluation method for multilingual systems, especially for minor-language systems.
Thus, as a first step, we develop retrieval-based conversational systems and evaluate the ability to select appropriate responses from a set of candidates.

\begin{figure}[t]
\begin{center}
\includegraphics[width=0.7\linewidth]{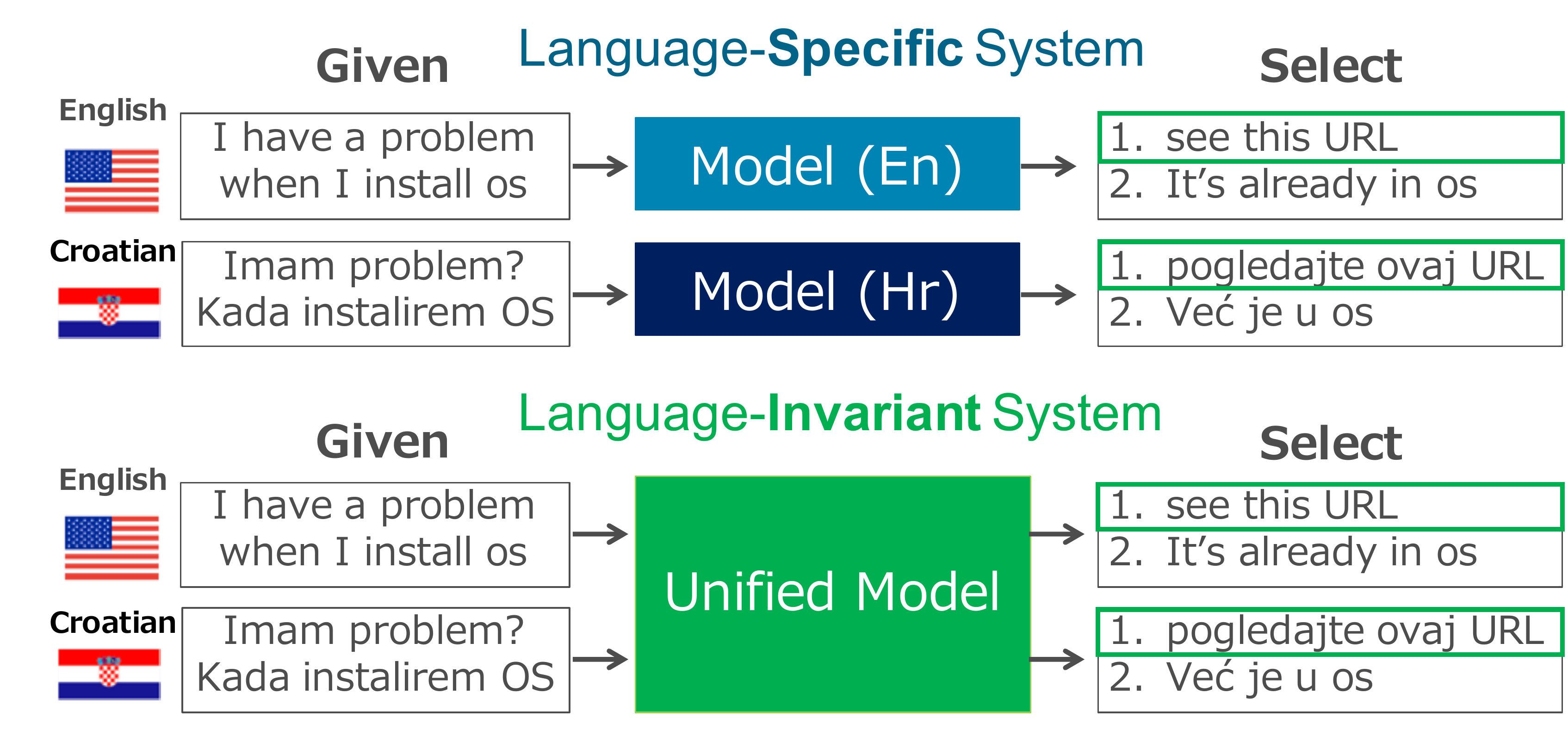}
\caption{Example of multilingual response selection.}
\label{fig:overview_of_task}
\end{center}
\end{figure}

Fig.\ \ref{fig:overview_of_task} shows the overview of our multilingual responding systems.

This paper provides: (i) formal task definitions, (ii) several knowledge-transfer methods and (iii) a multilingual conversation dataset.
First, we introduce and formalize the two task settings: {\it single-language adaptation} for language-specific systems and {\it multi-language adaptation} for language-invariant systems (Sec.\ \ref{sec:task}).

Second, we present several methods leveraging high-resource language data to compensate for low-resource language in the two settings (Sec.\ \ref{sec:method}).
Our basic method uses multilingual word embeddings and transfers source-language knowledge to target languages (Sec.\ \ref{sec:base} (a)).
We also design three extended methods.
Among them, the fine-tuning method builds a model specific to a single target language (Sec.\ \ref{sec:ext} (b)).
The joint loss training and the multi-language adversarial training methods build a unified model invariant for multiple target languages (Sec.\ \ref{sec:ext} (c) and (d)).

Third, we create a multilingual conversation corpus and dataset\footnote{Our code and dataset are publicly available at \url{https://github.com/aonotas/multilingual_ASR}} (Sec.\ \ref{sec:data}).

From the Ubuntu IRC Logs\footnote{http://irclogs.ubuntu.com/}, we collect the logs in $12$ languages.

To show benchmark results, we perform experiments on the created dataset (Sec.\ \ref{sec:exp}).
The results demonstrate that our methods allow models to effectively adapt to low-resource target languages.

In particular, our method using Wasserstein GAN \cite{arjovsky2017wasserstein} achieves high-performance for simultaneously dealing with multiple languages with a single unified model.

\section{Related Work}
\label{sec:rwork}

\subsection*{Short Text Conversation}
In short text conversation, a system predicts an appropriate response for an input message in {\it single-turn, two-party conversation} \cite{ritter:2011}.
One major approach to it is the generation-based approach, which generates a response using a sequence-to-sequence model \cite{shang:2015,vinyals:2015,serban:2016,li:2016,mei:2017:res}.
Another popular approach is the retrieval-based approach, which retrieves candidate responses from a repository and returns the highest scoring one using a ranking model \cite{wang:2013,lu:2013,ji:2014,wang:2015}.
\newcite{lowe:2015} proposed next utterance classification (NUC), in which a model has to select an appropriate response from a fixed set of candidates.

\subsection*{Evaluation for Conversational Systems}
Evaluation methods for conversational systems are an open question \cite{lowe:2015,liu:16,lowe:17}.

While many of previous studies on conversational systems used human judgements, automatic evaluation methods are attractive because it is much easier and less costly to use.
However, according to \newcite{liu:16}, for generation-based systems, automatic evaluation metrics, such as BLEU \cite{Papineni2002BleuAM}, correlate very weakly with human judgements.

For retrieval-based systems, some studies used ranking-based metrics, such as mean average precision and accuracy \cite{ji:2014,wang:2015}.
\newcite{Lowe2016OnTE} confirmed the feasibility of NUC as a surrogate task for building conversational systems.
Although there are controversial issues for these evaluation methods \cite{lowe:2015}, as a practical choice, we adopt the accuracy-based metric for evaluating multilingual conversational systems.

\subsection*{Addressee and Response Selection}
NUC focuses on two-party, {\it multi-turn conversation}.
As an extension of it, \newcite{ouchi-tsuboi:2016:EMNLP2016} proposed addressee and response selection (ARS) for {\it multi-party conversation}.
ARS integrates the addressee detection problem, which has been regarded as a problematic issue in multi-party conversation \cite{traum:2003,jovanovic:2004,Bohus:2011:MTT:2132890.2132903,uthus:2013}.
Mainly, this problem has been tackled in spoken/multimodal dialog systems \cite{jovanovic:2006,akker:2009,Nakano:2013:IEM:2522848.2522872,ravuri:2014}.
While these systems largely rely on acoustic signal or gaze information,
ARS focuses on text-based conversations.
Extending these studies, we tackle multilingual, multi-turn, and multi-party text conversation settings.

\subsection*{Cross-Lingual Conversation}
The motivation of our task is similar with that of \newcite{kim:2016:fifth}.
They tackled cross-lingual dialog state tracking in English and Chinese.
While they transfer knowledge from English to Chinese, we transfer knowledge between a high-resource and several low-resource languages.

\section{Addressee and Response Selection}
\label{sec:ars}
Addressee and response selection (ARS)\footnote{Due to the space limitation, we give a brief overview of ARS. For the complete task definition, please refer to \newcite{ouchi-tsuboi:2016:EMNLP2016}.}, proposed by \newcite{ouchi-tsuboi:2016:EMNLP2016}, assumes the situation where a responding agent returns a response to an addressee following a conversational context.

Formally, given an input conversational situation $\bvec{x} \in X$, a system predicts $\bvec{y} \in Y$, which consists of an addressee $a$ and a response $\bvec{r}$:
\[
\textsc{Given}: \bvec{x} = (a_{\rm res}, \: \mathcal{C}, \: \mathcal{R}),  \hspace{0.5cm} \textsc{Predict}: \bvec{y} = (a, \bvec{r})
\]

\noindent
where $a_{\rm res}$ is a responding agent, $\mathcal{C}$ is a context (a sequence of previous utterances) and $\mathcal{R}$ is a set of candidate responses.
To predict an addressee $a$, we select an agent from a set of the agents appearing in a context $\mathcal{A}(\mathcal{C})$.
To predict a response $\bvec{r}$, we select a response from a set of candidate responses $\mathcal{R} = \{ \bvec{r}_1, \cdots, \bvec{r}_{|\mathcal{R}|} \}$.

This task evaluates accuracy on the three aspects: addressee-response pair selection ({\tt ADR-RES}),
addressee selection ({\tt ADR}), and response selection ({\tt RES}).
In {\tt ADR-RES}, we regard the answer as correct if both the addressee and response are correctly selected.
In {\tt ADR}/{\tt RES}, we regard the answer as correct if the addressee/response is correctly selected.

\section{Multilingual Addressee and Response Selection}
\label{sec:task}
As an extension of monolingual ARS, we propose \textit{multilingual addressee and response selection} (M-ARS).
In ARS, a system is given as input a set of candidate responses and a conversational context in a single language.
By contrast, in M-ARS, a system receives the inputs in one of multiple languages.
In the following, we first explain our motivation for tackling M-ARS and then describe the formal task definitions.

\subsection{Motivative Situations}
We assume the two multilingual conversational  situations:
\begin{itemize}
  \item You want to build {\it language-specific} systems, each of which responds in a single language.
  \item You want to build one {\it language-invariant} system, which responds in multiple languages.
\end{itemize}
\noindent
The first situation is that we build $K$ models, each of which is specialized for one of $K$ target languages.
The second one is that we build one unified model that can deal with all $K$ target languages.
Taking these situations into account, we present the corresponding two tasks: (i) \textit{single-language adaptation} and (ii) \textit{multi-language adaptation}.

\subsection{Task Overview}
The goal of \textit{single-language adaptation} is to develop and evaluate a language-specific ARS model for a single target language.
For example, using English, German and Italian training data, we build a model specialized for German conversation.
The goal of \textit{multi-language adaptation} is to develop and evaluate a language-invariant ARS model for multiple target languages.
For example, using English, German and Italian training data, we build a model that can respond to not only German but also Italian and English conversation.
In the following subsections, we formalize each of these tasks.

\subsection{Formal Task Definition}
We assume that we have conversation data in each of a set of languages $\mathcal K$.\\

\vspace{-0.2cm}
\noindent
\textbf{Training}\\
\noindent
In the training phase, a training dataset is given for each language $k \in {\mathcal K}$:
\begin{eqnarray*}
\begin{aligned}
& \mathcal{D}^{(k)}_{\rm train} = \{ \: \bvec{x}^{(k)}_i, \bvec{y}^{(k)}_i \: \}_{i=1}^{N^{(k)}}, \:\: k \in \mathcal K\\
& \mathcal{D}_{\rm train} = \bigcup_k \: \mathcal{D}^{(k)}_{\rm train}
\end{aligned}
\end{eqnarray*}

\noindent
where $\bvec{x}^{(k)}$ and $\bvec{y}^{(k)}$ are a conversational situation and the target output in language $k$, respectively.
We train a model $\mathcal{F}: \mathcal{X} \rightarrow \mathcal{Y}$ on these training samples.\\

\noindent
\textbf{Evaluation}\\
\noindent
In single-language adaptation, we evaluate a trained model for a single target language $t \in \mathcal K$.
The trained model receives an input of the target language, $\bvec{x}^{(t)} \sim \mathcal{D}^{(t)}_{\rm eval}$, and predicts $\hat{\bvec{y}}^{(t)}$.
As evaluation metrics, we use the three accuracies ({\tt ADR-RES}, {\tt ADR} and {\tt RES}) used in ARS (Sec.\ \ref{sec:ars}).

In multi-language adaptation, given evaluation datasets for all the languages $\mathcal K$, i.e., $\bigcup_k \mathcal{D}^{(k)}_{\rm eval}$, the trained model receives an input of each language $\bvec{x}^{(k)} \sim \mathcal{D}^{(k)}_{\rm eval}$ and predicts $\hat{\bvec{y}}^{(k)}$.
As evaluation metrics, we use macro average over all the languages: $\text{\tt ADR-RES} = \frac{\sum_k \: \text{\tt ADR-RES}^{(k)}}{|\mathcal{K}|}$.
{\tt ADR} and {\tt RES} are also computed in the same way.

\section{Methods}
\label{sec:method}

In this section, we firstly describe a model used for addressee and response selection, and then explain our proposed methods to train parameters of the model.

\begin{figure}[t]
  \begin{center}
    \begin{tabular}{c}

      \begin{minipage}{0.49\hsize}
        \begin{center}
          \includegraphics[width=0.9\linewidth]{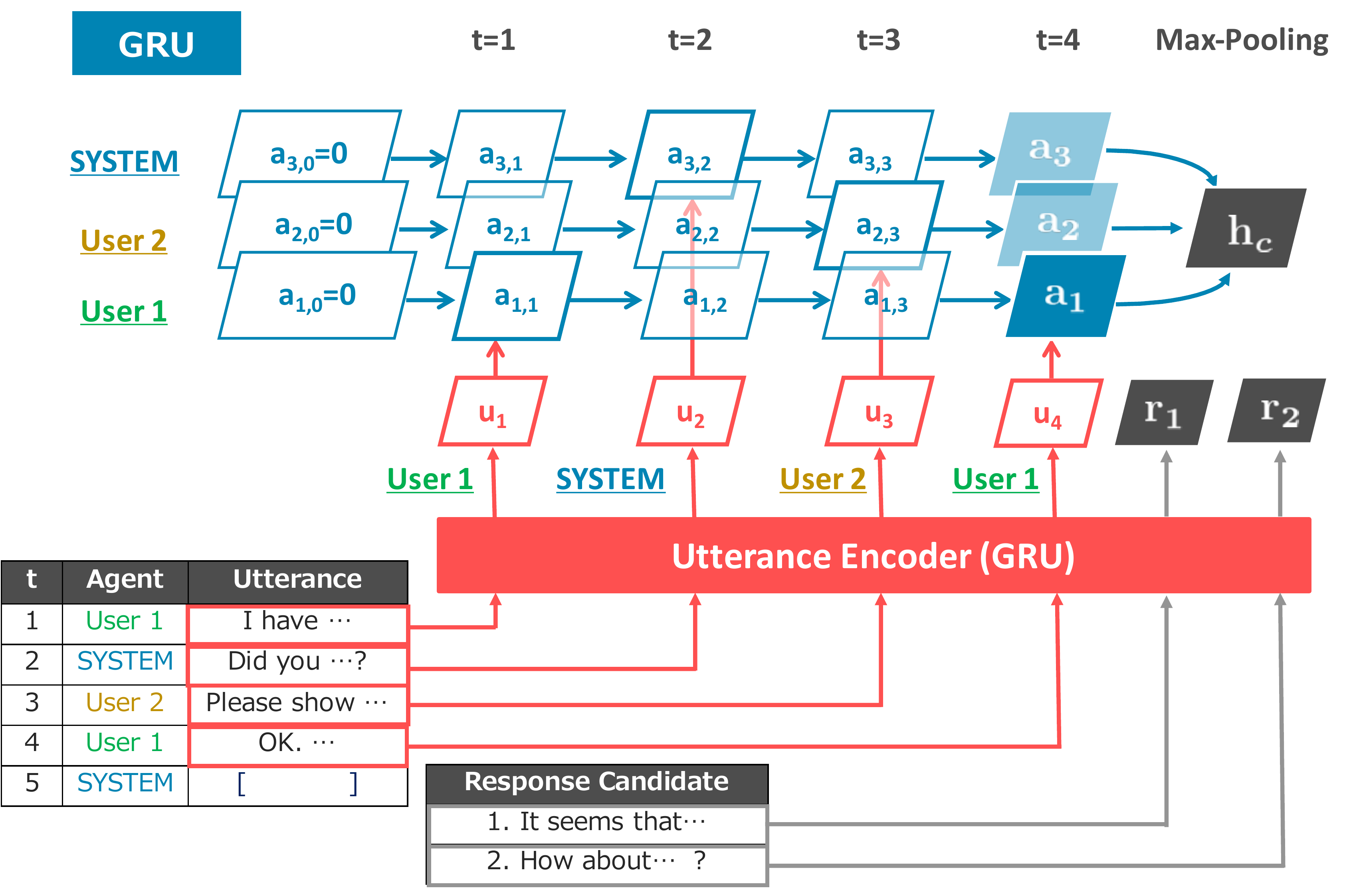}
          \caption{{Overview of Dynamic Model.}}
           \label{fig:overview_of_dynamic_model}
        \end{center}
      \end{minipage}

      \begin{minipage}{0.49\hsize}
        \begin{center}
          \includegraphics[width=0.9\linewidth]{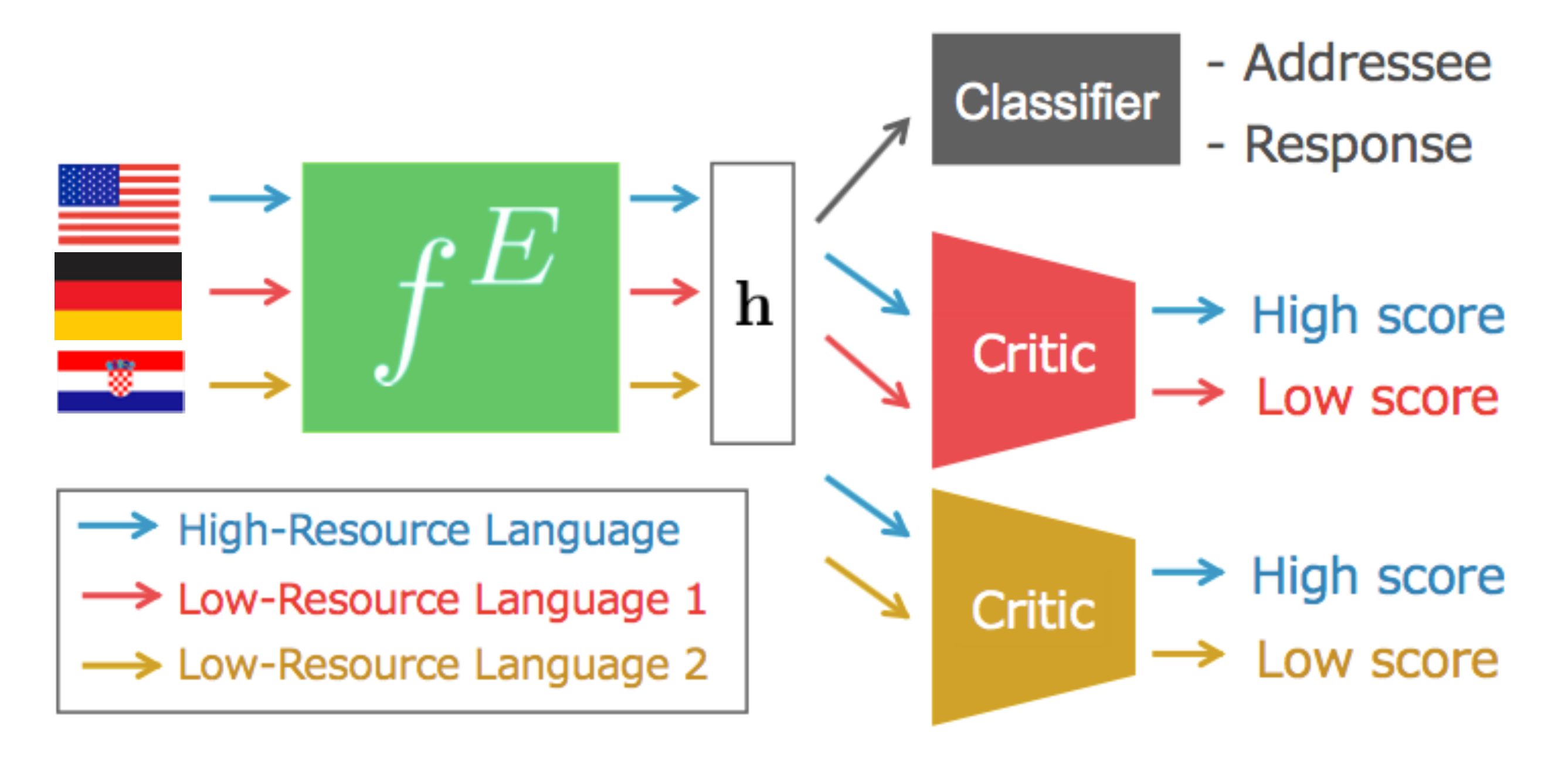}
          \caption{{Overview of our W-GAN training method for multiple target languages.}}
          \label{fig:overview_of_wgan}
        \end{center}
      \end{minipage}

    \end{tabular}
  \end{center}
\end{figure}

Our model $\mathcal{F}$ consists of a feature extractor $f^E$, addressee scoring function $f^A$ and response scoring function $f^R$.
$f^A$ and $f^R$ return relevance scores (probabilities) for an addressee and response:
\begin{eqnarray}
\label{eq:base1}
f^A(\bvec{x}, a_i) & = & \sigma([{\bf a}_{\rm res}, {\bf h}_c]^{\rm T} \: {\bf W}_a \: {\bf a}_i) \\
\label{eq:base2}
f^R(\bvec{x}, \bvec{r}_j)  & = & \sigma([{\bf a}_{\rm res}, {\bf h}_c]^{\rm T} \: {\bf W}_r \hspace{0.15cm} {\bf r}_j)
\end{eqnarray}

\noindent
where ${\bf a}_{\rm res}$ is a responding agent vector, ${\bf h}_c$ is a conversational context vector, ${\bf a}_i$ is an agent vector, and ${\bf r}_j$ is a candidate response vector.
All these vectors are encoded by the feature extractor $f^E$.
We use the dynamic model \cite{ouchi-tsuboi:2016:EMNLP2016} as $f^E$.
Fig.\ \ref{fig:overview_of_dynamic_model} shows the overview of the dynamic model.
This model represents each agent as a hidden state vector that dynamically changes along with time steps in GRU \cite{Cho2014LearningPR}.
\footnote{In this example, a responding agent vector ${\bf a}_{\rm res}$  is ${\bf a}_{3}$. Note that the states of the agents that are not speaking at the time are updated by zero vectors.}

A model $\mathcal{F}$ is parameterized by $\theta = \{ \theta_E \cup \{ {\bf W}_a, {\bf W}_r \} \}$, where $\theta_E$ is parameters of $f^E$.
To train these parameters, we present four methods.
These methods assume that we have training sets for a set of languages $\mathcal K$: some of them are high-resource languages $\mathcal{S} \subseteq \mathcal{K}$ and others are relatively low-resource languages $\mathcal{T} = \bar{\mathcal S}$.

\subsection{A Basic Method}
\label{sec:base}

\subsection*{(a) Multilingual Embedding Replacement}

This method trains a model $\mathcal{F}$ on high-resource language data $\mathcal{D}^{(s)}_{\rm train}$, where $s \in \mathcal S$, and uses the trained model for responding conversations in other languages $\bar{\mathcal S}$.
To realize this transfer, we use {\it multilingual embeddings}.

Consider the case where the high-resource language is English ({\tt En}) and low-resource language is German ({\tt De}).
In the training phase, we use English word embeddings to train model parameters.\footnote{The embeddings are fixed, not fine-tuned, during training.}
In the testing phase, instead of the English embeddings, we use German embeddings:\\

\noindent
\vspace{0.2cm}\hspace*{4cm}{\bf Train:} ${\bf w} = {\bf W}^{({\tt En})}_{\rm emb} \: w$ \hspace*{1cm}{\bf Test:} ${\bf w} = {\bf W}^{({\tt De})}_{\rm emb} \: w$\\
where $w$ is a one-hot vector.
We just replace the English word embeddings ${\bf W}^{({\tt En})}_{\rm emb}$ with the German ones ${\bf W}^{({\tt De})}_{\rm emb}$.
After looking up each word embedding ${\bf w}$, the neural model computes the hidden states.
One advantage of this method is to require no target language data.
As multilingual embeddings, we use MultiCCA\footnote{The pre-trained MultiCCA embeddings are provided at http://128.2.220.95/multilingual/data/} proposed by \newcite{DBLP:journals/corr/AmmarMTLDS16}.
In these embeddings, semantically similar words in the same language or translationally equivalent words in different languages are projected onto nearby.

Besides multilingual embeddings, another option to build a conversational model without no conversation data in a target language is to translate high-resource language data to low-resource one and train a conversational model on the translations.
One limitation of this approach is that it is costly to prepare parallel corpora for building the translation model.
We discuss this approach in Sec.\ \ref{sec:data_augmentation}.

\subsection{Extended Methods}
\label{sec:ext}
We present the two types of methods which use target language data for building (i) {\it language-specific models} and (ii) {\it language-invariant models}.

\subsubsection{Methods for Language-Specific Models}
\subsection*{(b) Fine-Tuning with Target Language Data}
\noindent
To compensate for the lack of the low-resource language data, this method firstly trains a model $\mathcal{F}_\theta$ on high-resource language data (pre-training phase).
Then, using the pre-trained parameters $\theta$ as the initial values, this method re-trains them on low-resource language data (fine-tuning phase).
We can expect that by gaining the better initial parameters, the tuning effectively adapts the model to the target language.

\subsubsection{Methods for Language-Invariant Models}
In order to build language-invariant models, it is critical to consider the two perspectives: (i) \textit{avoiding catastrophic forgetting} and (ii) \textit{learning language-invariant features}.
Catastrophic forgetting \cite{kirkpatrick2017overcoming} is the phenomenon that a model forgets knowledge of previously trained tasks (languages) by incorporating knowledge of the current task (language).
Language-invariant features are the features that are common and unchanged in different languages.
Taking these two perspectives into account, we present the following two methods.

\subsection*{(c) Joint Loss Training}
\noindent
This method aims to avoid catastrophic forgetting by jointly training model parameters on all the language data at a time.
Assuming that we have a set of languages $\mathcal K$, this method uses the joint loss function: $\mathcal{J}_{\text{joint}}(\theta) = \sum_k \mathcal{J}(\mathcal{D}^{(k)}, \theta)$
%
where the loss function $\mathcal{J}$ is the cross-entropy loss used in \newcite{ouchi-tsuboi:2016:EMNLP2016}.

\subsection*{(d) Multi-Language Adversarial Training}
\noindent
To learn language-invariant features, we use a framework of Wasserstein-GAN (W-GAN) \cite{arjovsky2017wasserstein}, a recently proposed technique to improve stability for generative adversarial nets (GANs) \cite{goodfellow2014generative}.
The aim of this method is to match the distributions of feature representations in two languages.

Fig.\ \ref{fig:overview_of_wgan} illustrates an example.
English is the high-resource language $s \in \mathcal S$, and German and Croatian are low-resource languages $t \in \mathcal T$.
For each language, the feature extractor $f^E$ receives an input conversation $\bvec{x}$ and computes the hidden features ${\bf h} = f^E(\bvec{x})$\footnote{Hidden feature representation $\bf h$ is the concatenation of the responding speaker vector and context vector in Eqs.\ \ref{eq:base1} and \ref{eq:base2}, i.e. ${\bf h} = [{\bf a}_{\rm res}, {\bf h}_c]$.}.
Thus, by using $f^E$, we obtain the hidden feature ${\bf h}^{(s)}$ and ${\bf h}^{(t)}$ for English and the others, respectively.

A pair of the high- and low-resource language features ${\bf h}^{(s)}$ and ${\bf h}^{(t)}$ is given to a critic $g_\pi$ to minimize the Wasserstein distance between the distributions $p({\bf h}^{(s)})$ and $p({\bf h}^{(t)})$:
\begin{eqnarray}
\label{eq:wdist}
 \mathcal{W}(\: p({\bf h}^{(s)}), \: p({\bf h}^{(t)}) \: ) = \max_\pi \: \mathbb E_{{\bf h}^{(s)}\sim p({\bf h}^{(s)})}[g_\pi({\bf h}^{(s)})] - \mathbb E_{{\bf h}^{(t)} \sim p({\bf h}^{(t)})}[g_\pi({\bf h}^{(t)})]
\end{eqnarray}

\noindent
where the maximum is taken over the set of all 1-Lipschitz functions $g_\pi$.\footnote{A function $g$ is 1-Lipschitz when $|g(x) - g(y)| \leq |x - y|$ for all $x$ and $y$. To constrain the critic $g$ to a 1-Lipschitz function, the parameters of $g$ are clipped to a fixed range.}
By maximizing this equation, the distributions of the feature representations, $p({\bf h}^{(s)})$ and $p({\bf h}^{(t)})$, are made as close as possible.
In this paper, as the critic $g_\pi$, we use multi-layer perceptron.

Eq.\ \ref{eq:wdist} is designed for the two distributions.
Thus, we generalize this W-GAN equation to deal with $|\mathcal{S}|$ high-resource languages and $|\mathcal{T}|$ low-resource languages:
\[
\mathcal{J}_{\text{wgan}}(\theta) = \sum_{s \in \mathcal{S}} \sum_{t \in \mathcal{T}} \: \mathcal{W}(\: p({\bf h}^{(k)}), \: p({\bf h}^{(\ell)}) \: )
\]
\noindent
This loss function $\mathcal{J}_{\text{wgan}}$ is integrated with the joint loss: $\mathcal{J}_{\text{adv}}(\theta) = \mathcal{J}_{\text{joint}}(\theta) + \lambda \:  \mathcal{J}_{\text{wgan}}(\theta)$
where $\lambda$ is a hyper-parameter that balances these loss functions and we used $\lambda=0.5$.

\begin{table}[t]
  \begin{minipage}[t]{.5\textwidth}

    {\small
      \begin{center}
      \begin{tabular}{lrrr} \toprule
        \multicolumn{4}{c}{Corpus} \\
        Language          & Docs  &  Utters &  Words  \\ \midrule
        English (en)      & 7355  &   2.4 M &  27.0 M \\
        Italian (it)      &  357  &   165 k &   1.1 M \\
        Croatian (hr)     &  254  &    80 k &   630 k \\
        German (de)       &  248  &    38 k &   335 k \\
        Portuguese (pt)   &  211  &    52 k &   285 k \\
        Slovenian (sl)    &  179  &    59 k &   357 k \\
        Polish(pl)        &   67  &   8.8 k &    51 k \\
        Dutch (nl)        &   57  &   7.2 k &    75 k \\
        Spanish (es)      &   36  &   7.1 k &    49 k \\
        Swedish (sv)      &   26  &   1.7 k &   6.8 k \\
        Russian (ru)      &    5  &   0.3 k &   1.5 k \\
        French (fr)       &    3  &   0.5 k &   3.0 k \\
        \bottomrule
      \end{tabular}
      \end{center}
      \caption{{Statistics of M-ARS corpus.}}
      \label{tab:oter_lang_corpus}
      }
  \end{minipage}
  \hfill
  \begin{minipage}[t]{.5\textwidth}

    {\small
      \begin{center}
      \begin{tabular}{lrrr} \toprule
        \multicolumn{4}{c}{Dataset} \\
        Language          & Train   &  Dev   &  Test \\ \midrule
        English (en)      & 665.6 k & 45.1 k & 51.9 k  \\
        Italian (it)      & 38,511  & 2,561  &  3,873  \\
        Croatian (hr)     & 11,387  &   512  &  1,145  \\
        German (de)       &  5,500  &   354  &  569  \\
        Portuguese (pt)   &  5,951  &   285  &  975  \\
        \bottomrule
      \end{tabular}
      \end{center}
      \caption{\label{tab:dataset}
      {
        Statistics of the M-ARS dataset.
      }}
      }

  \end{minipage}
\end{table}

\section{Corpus and Dataset}
\label{sec:data}
One of our goals is to provide a multilingual conversation corpus/dataset that can be used over a wide range of conversation research.
We follow the corpus and data creation method of \newcite{ouchi-tsuboi:2016:EMNLP2016}.
First, we crawl the Ubuntu IRC Logs\footnote{We use a collection of the logs during one year (2015). We plan to expand it by collecting the logs over all the years.},
and preprocess the logs in many languages.
Each language is identified by using a language detection library \cite{nakatani:2010}.
The resulting corpus consists of multilingual conversations in $12$ languages, shown in Tab.\ \ref{tab:oter_lang_corpus}.

Then, we create an M-ARS dataset.
For each language, we set the ground-truth/false addressees and responses following \newcite{ouchi-tsuboi:2016:EMNLP2016}.
Note that the addressed usernames in utterances have been removed for addressee selection. Thus, we have to predict the addressees without seeing the addressed usernames.
The number of candidate responses ($|\mathcal{R}|$) is set to $2$ or $10$.
The dataset is then randomly partitioned into a training set (90\%), a development set (5\%) and a test set (5\%).
Tab.\ \ref{tab:dataset} shows the statistics of the top $5$ largest language sections of this resulting dataset.

\begin{table*}[t]
  {\small
  \begin{center}
  \begin{tabular}{c|l|ccc|ccc} \toprule
    & & \multicolumn{3}{c|}{$|\mathcal{R}|=2$} & \multicolumn{3}{c}{$|\mathcal{R}|=10$} \\
    Setting &Method & {\tt ADR-RES} & {\tt ADR} & {\tt RES} & {\tt ADR-RES} & {\tt ADR} & {\tt RES} \\ \midrule
    \multirow{7}{*}{Single Language Adaptation}
      & \sc Chance               & \,\,\,3.97	&  \,\,\,7.94	& 50.00	& \,\,\,0.80	& \,\,\,7.94	& 10.00 \\
      & \sc TF-IDF               & 39.51	& 64.97	& 60.61	& 12.54	& 64.97	& 18.50 \\
      & \sc TrgOnly              & 47.35	& 69.27	& 67.35	& 19.42	& 69.73	& 26.13 \\
      & \sc EnOnly               & 38.07	& 65.72	& 57.65	& \,\,\,8.50	& 62.38	& 13.75 \\
      & \sc FineTune             & 49.58	& 69.59	& 69.84	& 21.15	& 70.33	& 28.15 \\
      & \sc Joint                & 51.55	& 70.30	& 71.88	& 22.32	& \bf 70.36	& 29.38 \\
      & \sc Wgan                 & \bf 53.17	& \bf 70.99	& \bf 73.25	& \bf 23.34	& 70.20	& \bf 30.39 \\ \midrule 
      \multirow{6}{*}{Two Language Adaptation}
       & \sc Chance              & \,\,\,2.30	 &  \,\,\,4.59 &  50.00 & \,\,\,0.46  &  	\,\,\,4.59 & 	10.00\\
       & \sc TF-IDF              & 38.32 & 60.29 &  64.25 & 13.99 &  60.29 &  23.84  \\
       & \sc EnOnly              & 46.77 & 67.62 & 67.90 & 19.88 & 65.86 & 27.83 \\
       & \sc FineTune            & 50.98 & 68.79 & 72.60 & 24.30 & 68.89 & 32.96\\
       & \sc Joint               & 53.37 & 69.75 & 74.94 & 26.60 & 69.75 & 35.59\\
       & \sc Wgan                & \bf 54.14 & \bf 70.07 & \bf 75.63 & \bf 27.23 & \bf 69.76 & \bf 36.11 \\ \midrule 
      \multirow{4}{*}{Five Language Adaptation}
      & \sc TF-IDF              & 39.04 & 63.10 & 62.06 & 13.12 & 63.10 & 20.64 \\
      & \sc EnOnly              & 41.55 & 66.48 & 61.75 & 13.05 & 63.77 & 19.38 \\
      & \sc Joint               & 50.69 & 69.00 & 72.18 & 22.80 & 69.18 & 31.11 \\
      & \sc Wgan                & \bf 52.11 & \bf 69.74 & \bf 73.34 & \bf 23.39 & \bf 69.35 & \bf 31.88 \\
     \bottomrule
  \end{tabular}
  \end{center}}
  \caption{\label{tab:result} Results for Single-/Two-/Five- language adaptation. Each number represents accuracy on addressee-response selection ({\tt ADR-RES}), addressee selection ({\tt ADR}) or response selection ({\tt RES}). 
  }
\end{table*}
\section{Experiments}
\label{sec:exp}

\subsection{Experimental Setup}
\subsubsection{Task Settings}
We use the following languages: (i) English ({\tt En}) as the high-resource language, and (ii) Italian ({\tt It}), Croatian ({\tt Hr}), German ({\tt De}), Portuguese ({\tt Pt}) as the low-resource languages.
In the following, we describe the languages used in each task.

\subsection*{(a) Single-Language Adaptation}
\begin{description}
\item[] \hspace{0.3cm} {\bf Train:} English + 1 Low-Res. Language,  \,\,\,\,\,\,\,\,\,\,\,\,\,\,{\bf Dev \& Test:} 1 Low-Res. Language \vspace{-0.2cm}
\end{description}
\noindent
For example, in the case that the target is Italian ({\tt It}), we use the {\tt En} and {\tt It} training sets to train a model, and evaluate the trained model on the {\tt It} test set.
As evaluation metrics, we use the three types of accuracies, {\tt ADR-RES}, {\tt ARD} and {\tt RES} (described in Sec.\ \ref{sec:ars}).
We report the macro average accuracies of all source-target language pairs ({\tt En}-{\tt It}, {\tt En}-{\tt Hr}, {\tt En}-{\tt De}, and {\tt En}-{\tt Pt}).

\subsection*{(b) Multi-Language Adaptation}
\begin{description}
\item[] \hspace{0.3cm} {\bf Train:}  English + $|\mathcal{T}|$ Low-Res. Languages, \,\,\,\,\,\,{\bf Dev \& Test:} English + $|\mathcal{T}|$ Low-Res. Languages \vspace{-0.2cm}
\end{description}
\noindent
We use the {\tt En}, {\tt It}, {\tt Hr}, {\tt Pt} and {\tt De} training sets to train a unified model, and evaluate it on the test sets for all the languages ({\tt En}, {\tt It}, {\tt Hr}, {\tt Pt}, {\tt De}).
As evaluation metrics, we use the macro averages of {\tt ADR-RES}, {\tt ARD} and {\tt RES} for all the languages.
For example, for two language adaptation ($|\mathcal{T}|=1$),
we report the macro averages over all the language pairs ({\tt En}-{\tt It}, {\tt En}-{\tt Hr}, {\tt En}-{\tt De}, and {\tt En}-{\tt Pt}).
For five language adaptation ($|\mathcal{T}|=4$),
we report the macro averages over all the five languages.
Note that while we evaluate the performance on only the test set of the target low-resource language in single-language adaptation, we evaluate it on the test sets of English and the low-resource languages in multi-language adaptation.

\subsubsection{Comparative Methods}
We compare several methods.
Our proposed methods (Sec.\ \ref{sec:method}) are orthogonal, so that we can combine a method with others.
In the following, we list the methods used in the comparison.
\begin{itemize}
  \setlength{\itemsep}{0.15cm}
  \item {\sc TrgOnly}: A dynamic model proposed by \newcite{ouchi-tsuboi:2016:EMNLP2016} trained on only the low-resource target language data.\vspace{-0.2cm}
  \item {\sc EnOnly}: A model built by (a) {\it multilingual embedding replacement} in Sec.\ \ref{sec:base}: training a model on the English data and replacing the English word embeddings with the embeddings of the low-resource language.\vspace{-0.2cm}
  \item {\sc FineTune}: A model built by (b) {\it fine-tuning} in Sec.\ \ref{sec:ext}: training a model on the high-resource language (English), and retraining it on the low-resource language.\vspace{-0.2cm}
  \item {\sc Joint}: A model built by (b) {\it fine-tuning} and (c) {\it joint loss training}: building a model by {\sc FineTune} as an initial model, and retraining it with the joint loss functions.\vspace{-0.2cm}
  \item {\sc Wgan}: A model built by (b) {\it fine-tuning} and (d) {\it multi-language adversarial training}: building a model by {\sc FineTune} as an initial model, and retraining it with W-GAN.
\end{itemize}

\noindent
Besides the neural models, we also use the {\sc TF-IDF} model used in \newcite{ouchi-tsuboi:2016:EMNLP2016}.
This model firstly creates TF-IDF vectors for the context and each candidate response.
Then, it computes the cosine similarity for each pair of the context vector and a response vector.
Finally, it selects the candidate response with the highest similarity.

\subsubsection{Optimization}
We use stochastic gradient descent (SGD) with a mini-batch method.
To update parameters, we use \textit{Adam} \cite{kingma:2014}.
We describe the details of hyper-parameter settings in Supplementary Material.

\subsection{Results}
Tab.\ \ref{tab:result} shows the results of single-language and multi-language adaptation.
Note that Tab. \ref{tab:result_single}, Tab. \ref{tab:result_two}, and Tab. \ref{tab:result_five} shows the detailed results for each language.

\subsubsection*{Single-Language Adaptation}
{\sc Wgan} achieved the best scores for most of the metrics.
This suggests that the W-GAN method successfully transfers knowledge of high-resource language to a low-resource language.
Also, {\sc FineTune} outperformed {\sc TrgOnly}.
This means that pre-training parameters on the high-resource language data improves a model for a target low-resource language.
Interestingly, {\sc EnOnly} achieved higher scores than chance-level without any target language data.
One possible explanation is that the multilingual embeddings have good alignments to some extent between similar meaning words in different languages.


\subsubsection*{Multi-Language Adaptation}
In both two- and five-language adaptation, {\sc Wgan} achieved the best scores.
Specifically, In five language adaptation, regardless of using a single, unified model, {\sc Joint} and {\sc Wgan} achieved high-performance.\footnote{Since {\sc FineTune} builds a model for each target language, it cannot analyze multiple languages with a single model. That is why there is no result of {\sc FineTune} in five-language adaptation.}

Also, {\sc Wgan} outperformed {\sc Joint} in all the metrics.
This suggests that {\sc Wgan} learns language-invariant features more effectively.

In NLP tasks, \newcite{chen2016adversarial} applied W-GAN to cross-lingual sentiment classification and successfully transferred the source-side knowledge to the target one.
In this paper, we have extended the adaptation of single source-target pair to the adaptation of multiple pairs. 
Our experimental results show that our method works well for multi-language adaptation in conversation domain.

\subsection{Data Augmentation with NMT}
\label{sec:data_augmentation}
\begin{figure}[t]
  \begin{center}
    \begin{tabular}{c}

      \begin{minipage}{0.49\hsize}
        \begin{center}
          \includegraphics[width=0.9\linewidth]{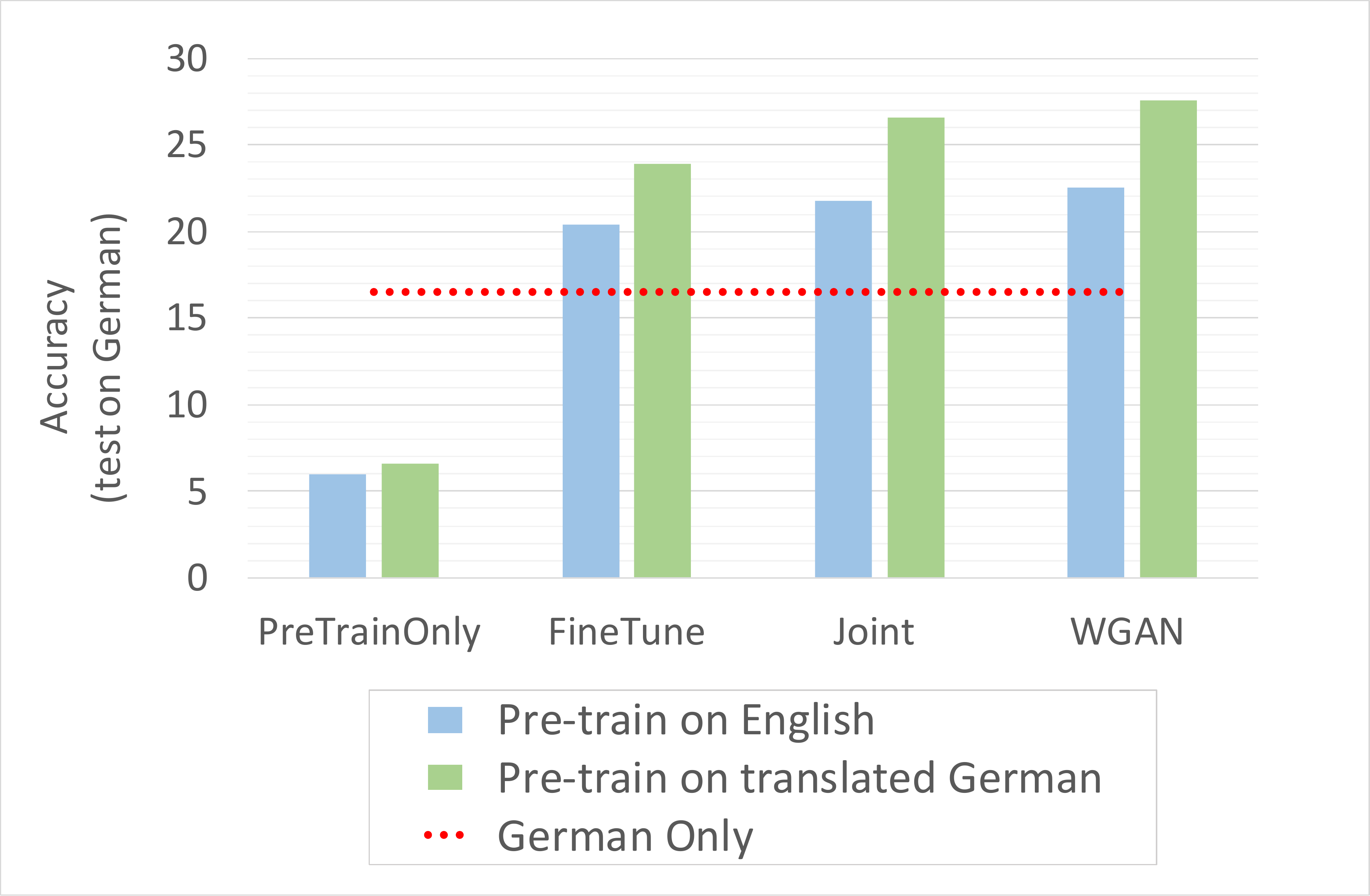}
          \caption{{\small Effects of data augmentation with NMT.}}
           \label{fig:plot_data_augmentation}
        \end{center}
      \end{minipage}

      \begin{minipage}{0.49\hsize}
        \begin{center}
          \includegraphics[width=0.9\linewidth]{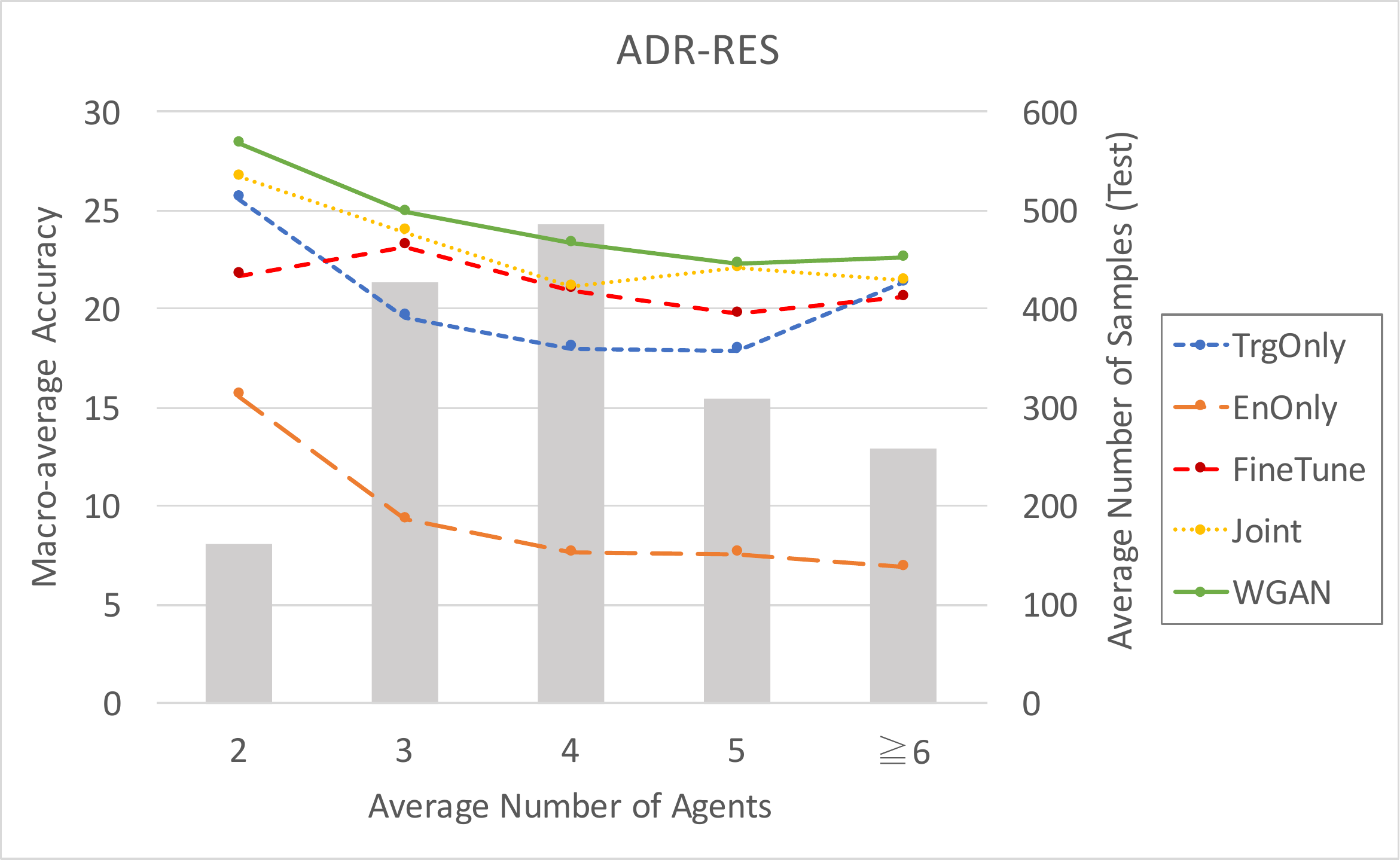}
            \caption{{\small Effects of the number of agents in the context. Left axis: {\tt ADR-RES} accuracy on test sets (drawn as lines). Right axis: the average number of test samples (drawn as bars). }}
          \label{fig:analysis_plot}
        \end{center}
      \end{minipage}

    \end{tabular}
  \end{center}
\end{figure}

As we mentioned in Sec. \ref{sec:base} (a), as another approach to compensating for low-resource language, we use data augmentation.
To increase the amount of the training set of a low-resource language, we translate high-resource language (English) samples to low-resource ones by using Neural Machine Translation (NMT).
Although some translations are noisy, we can obtain much more training samples for a low-resource language.
One limitation of this method is that it is costly to prepare parallel corpora, which is often unavailable for low-resource languages.
For reproducibility, we use publicly available NMT models already trained on a parallel corpus.
Since OpenNMT\footnote{http://opennmt.net/Models/} provides an English-German model, we conduct experiments for the English-German pair.

We investigate the effects of translated German data for pre-training a model.
Translating English ({\tt En}) training utterances to German ({\tt De}) ones by using the trained NMT model, we can obtain translated German training data ({\tt De'}).
We compare the two settings: (i) pre-training a model on English data and (ii) pre-training a model on translated German data.
After pre-training, we apply the methods used in Sec.\ \ref{sec:exp}.
We evaluate the performance with {\tt ADR-RES} accuracy on the German test set.

Fig.\ \ref{fig:plot_data_augmentation} shows the results ($|\mathcal{R}|=10$).
The red dotted line is the performance of the modes trained on only original German training data ({\tt De}).
In each method, the blue bar at left hand is a model pre-trained on English ({\tt En}), and the green bar at right hand is a model pre-trained on ({\tt De'}).

{\sc PreTrainOnly}, the left-most method in Fig.\ref{fig:plot_data_augmentation}, uses the pre-trained models.
The results were almost the same between models pre-trained on English or translated German data, and worse than the model trained on only the original German training data (red dotted line).
This suggests that only the translations by NMT are not sufficient for building good multilingual ARS models.

Furthermore, we re-train the pre-trained models by using the three methods, {\sc FineTune}, {\sc Joint} and {\sc Wgan}.
In other words, each method uses a pre-trained model as the initial model and re-trains the parameters.
In all methods, the models re-trained from the {\tt De'} pre-trained model (green bars) were better than the ones from the {\tt En} pre-trained ones (blue bars).
This suggests that by combining the NMT-based data augmentation method with the knowledge-transfer methods, the performance is boosted.
Another point is that {\sc Wgan} consistently outperformed the other methods, which supports the utility of {\sc Wgan}.

\subsection{Analysis of Number of Agents}
\label{sec:model_analysis}
We investigate how accuracy fluctuates according to the number of agents in the context of length $15$, as shown in Fig. \ref{fig:analysis_plot}.
Overall, as the number of agents increases, the accuracies of all the methods tend to decline.
Among them, {\sc Wgan} achieved the best results in most of the cases.
This suggests that {\sc Wgan} can stably predicts addressees and responses in conversations with many participants.

\section{Conclusion and Future Work}
\label{sec:conc}
We have introduced {\it multilingual addressee and response selection} by providing
(i) formal task definitions, (ii) several knowledge-transfer methods and (iii) a multilingual conversation corpus and dataset.

Experimental results have demonstrated that our methods allow models to adapt multiple target languages.
In particular, methods for language-invariant models can simultaneously deal with multiple languages with a single model.

Since our methods and dataset can apply to response generation, tackling the multilingual response generation tasks is an interesting line of our future work.
In addition, our language-invariant systems can receive conversation in a language (e.g., English ) and reply to it in another language (e.g., German).
It can lead to interesting findings that our system is evaluated on code-mixing situations, where two or more languages are used in the same context.

\bibliographystyle{acl}
\bibliography{coling2018}

\begin{thebibliography}{}

\bibitem[\protect\citename{Akker and Traum}2009]{akker:2009}
Rieks Akker and David Traum.
\newblock 2009.
\newblock A comparison of addressee detection methods for multiparty
  conversations.
\newblock In {\em Workshop on the Semantics and Pragmatics of Dialogue}.

\bibitem[\protect\citename{Ammar \bgroup et al.\egroup
  }2016]{DBLP:journals/corr/AmmarMTLDS16}
Waleed Ammar, George Mulcaire, Yulia Tsvetkov, Guillaume Lample, Chris Dyer,
  and Noah~A. Smith.
\newblock 2016.
\newblock Massively multilingual word embeddings.
\newblock {\em CoRR}.

\bibitem[\protect\citename{Arjovsky \bgroup et al.\egroup
  }2017]{arjovsky2017wasserstein}
Martin Arjovsky, Soumith Chintala, and L{\'e}on Bottou.
\newblock 2017.
\newblock Wasserstein gan.
\newblock \textit{arXiv preprint arXiv:1701.07875}.

\bibitem[\protect\citename{Bohus and
  Horvitz}2011]{Bohus:2011:MTT:2132890.2132903}
Dan Bohus and Eric Horvitz.
\newblock 2011.
\newblock Multiparty turn taking in situated dialog: Study, lessons, and
  directions.
\newblock In {\em Proceedings of the SIGDIAL 2011 Conference}, SIGDIAL '11,
  pages 98--109, Stroudsburg, PA, USA. Association for Computational
  Linguistics.

\bibitem[\protect\citename{Chen \bgroup et al.\egroup
  }2016]{chen2016adversarial}
Xilun Chen, Yu~Sun, Ben Athiwaratkun, Claire Cardie, and Kilian Weinberger.
\newblock 2016.
\newblock Adversarial deep averaging networks for cross-lingual sentiment
  classification.

\bibitem[\protect\citename{Cho \bgroup et al.\egroup }2014]{Cho2014LearningPR}
Kyunghyun Cho, Bart van Merrienboer, Çaglar G{\"u}lçehre, Dzmitry Bahdanau,
  Fethi Bougares, Holger Schwenk, and Yoshua Bengio.
\newblock 2014.
\newblock Learning phrase representations using rnn encoder-decoder for
  statistical machine translation.
\newblock In {\em EMNLP}.

\bibitem[\protect\citename{Goodfellow \bgroup et al.\egroup
  }2014]{goodfellow2014generative}
Ian Goodfellow, Jean Pouget-Abadie, Mehdi Mirza, Bing Xu, David Warde-Farley,
  Sherjil Ozair, Aaron Courville, and Yoshua Bengio.
\newblock 2014.
\newblock Generative adversarial nets.
\newblock In {\em Proceedings of NIPS}, pages 2672--2680.

\bibitem[\protect\citename{Ji \bgroup et al.\egroup }2014]{ji:2014}
Zongcheng Ji, Zhengdong Lu, and Hang Li.
\newblock 2014.
\newblock An information retrieval approach to short text conversation.
\newblock \textit{arXiv preprint arXiv: 1408.6988}.

\bibitem[\protect\citename{Jovanovi{\'c} and Akker}2004]{jovanovic:2004}
Natasa Jovanovi{\'c} and op~den~Rieks Akker.
\newblock 2004.
\newblock Towards automatic addressee identification in multi-party dialogues.
\newblock In {\em Proceedings of SIGDIAL}.

\bibitem[\protect\citename{Jovanovi{\'c} \bgroup et al.\egroup
  }2006]{jovanovic:2006}
Natasa Jovanovi{\'c}, op~den~Rieks Akker, and Anton Nijholt.
\newblock 2006.
\newblock Addressee identification in face-to-face meetings.
\newblock In {\em Proceedings of EACL}.

\bibitem[\protect\citename{Kim \bgroup et al.\egroup }2016]{kim:2016:fifth}
Seokhwan Kim, Luis~Fernando D'Haro, Rafael~E Banchs, Jason~D Williams, Matthew
  Henderson, and Koichiro Yoshino.
\newblock 2016.
\newblock The fifth dialog state tracking challenge.
\newblock In {\em Proceeding of Spoken Language Technology Workshop (SLT)},
  pages 511--517.

\bibitem[\protect\citename{Kingma and Ba}2014]{kingma:2014}
Diederik~P. Kingma and Jimmy~Lei Ba.
\newblock 2014.
\newblock Adam: A method for stochastic optimization.
\newblock \textit{arXiv preprint arXiv: 1412.6980}.

\bibitem[\protect\citename{Kirkpatrick \bgroup et al.\egroup
  }2017]{kirkpatrick2017overcoming}
James Kirkpatrick, Razvan Pascanu, Neil Rabinowitz, Joel Veness, Guillaume
  Desjardins, Andrei~A Rusu, Kieran Milan, John Quan, Tiago Ramalho, Agnieszka
  Grabska-Barwinska, et~al.
\newblock 2017.
\newblock Overcoming catastrophic forgetting in neural networks.
\newblock {\em Proceedings of the National Academy of Sciences of the United
  States of America}, pages 3521--3526.

\bibitem[\protect\citename{Li \bgroup et al.\egroup }2016]{li:2016}
Jiwei Li, Michel Galley, Chris Brockett, Jianfeng Gao, and Bill Dolan.
\newblock 2016.
\newblock A persona-based neural conversation model.
\newblock In {\em Proceedings of ACL}.

\bibitem[\protect\citename{Liu \bgroup et al.\egroup }2016]{liu:16}
Chia-Wei Liu, Ryan Lowe, Iulian Serban, Mike Noseworthy, Laurent Charlin, and
  Joelle Pineau.
\newblock 2016.
\newblock How not to evaluate your dialogue system: An empirical study of
  unsupervised evaluation metrics for dialogue response generation.
\newblock In {\em Proceedings of EMNLP}, pages 2122--2132.

\bibitem[\protect\citename{Lowe \bgroup et al.\egroup }2015]{lowe:2015}
Ryan Lowe, Nissan Pow, Iulian V.~Serban, and Joelle Pineau.
\newblock 2015.
\newblock The ubuntu dialogue corpus: A large dataset for research in
  unstructured multi-turn dialogue systems.
\newblock In {\em Proceedings of SIGDIAL}, pages 285--294.

\bibitem[\protect\citename{Lowe \bgroup et al.\egroup }2016]{Lowe2016OnTE}
Ryan Lowe, Iulian~Vlad Serban, Michael Noseworthy, Laurent Charlin, and Joelle
  Pineau.
\newblock 2016.
\newblock On the evaluation of dialogue systems with next utterance
  classification.
\newblock In {\em Proceedings of the 17th Annual Meeting of the Special
  Interest Group on Discourse and Dialogue}, pages 264--269, Los Angeles,
  September. Association for Computational Linguistics.

\bibitem[\protect\citename{Lowe \bgroup et al.\egroup }2017]{lowe:17}
Ryan Lowe, Michael Noseworthy, Iulian~Vlad Serban, Nicolas Angelard-Gontier,
  Yoshua Bengio, and Joelle Pineau.
\newblock 2017.
\newblock Towards an automatic turing test: Learning to evaluate dialogue
  responses.
\newblock In {\em Proceedings of ACL}, pages 1116--1126.

\bibitem[\protect\citename{Lu and Li}2013]{lu:2013}
Zhengdong Lu and Hang Li.
\newblock 2013.
\newblock A deep architecture for matching short texts.
\newblock In {\em Proceedings of NIPS}, pages 1367--1375.

\bibitem[\protect\citename{Mei \bgroup et al.\egroup }2017]{mei:2017:res}
Hongyuan Mei, Mohit Bansal, and Matthew~R Walter.
\newblock 2017.
\newblock Coherent dialogue with attention-based language models.
\newblock In {\em Proceedings of AAAI}.

\bibitem[\protect\citename{Nakano \bgroup et al.\egroup
  }2013]{Nakano:2013:IEM:2522848.2522872}
Yukiko~I. Nakano, Naoya Baba, Hung-Hsuan Huang, and Yuki Hayashi.
\newblock 2013.
\newblock Implementation and evaluation of a multimodal addressee
  identification mechanism for multiparty conversation systems.
\newblock In {\em Proceedings of the 15th ACM on International Conference on
  Multimodal Interaction}, ICMI '13, pages 35--42, New York, NY, USA. ACM.

\bibitem[\protect\citename{Nakatani}2010]{nakatani:2010}
Shuyo Nakatani.
\newblock 2010.
\newblock Language detection library for java.

\bibitem[\protect\citename{Ouchi and Tsuboi}2016]{ouchi-tsuboi:2016:EMNLP2016}
Hiroki Ouchi and Yuta Tsuboi.
\newblock 2016.
\newblock Addressee and response selection for multi-party conversation.
\newblock In {\em Proceedings of EMNLP}, pages 2133--2143.

\bibitem[\protect\citename{Papineni \bgroup et al.\egroup
  }2002]{Papineni2002BleuAM}
Kishore Papineni, Salim~E. Roucos, Todd Ward, and Wei-Jing Zhu.
\newblock 2002.
\newblock Bleu: a method for automatic evaluation of machine translation.
\newblock In {\em ACL}.

\bibitem[\protect\citename{Ravuri and Stolcke}2014]{ravuri:2014}
Suman~V Ravuri and Andreas Stolcke.
\newblock 2014.
\newblock Neural network models for lexical addressee detection.
\newblock In {\em Proceedings of INTERSPEECH}, pages 298--302.

\bibitem[\protect\citename{Ritter \bgroup et al.\egroup }2011]{ritter:2011}
Alan Ritter, Colin Cherry, and William~B. Dolan.
\newblock 2011.
\newblock Data-driven response generation in social media.
\newblock In {\em Proceedings of EMNL}, pages 583--593.

\bibitem[\protect\citename{Serban \bgroup et al.\egroup }2016]{serban:2016}
Iulian~Vlad Serban, Alessandro Sordoni, Yoshua Bengio, Aaron Courville, and
  Joelle Pineau.
\newblock 2016.
\newblock Building end-to-end dialogue systems using generative hierarchical
  neural network models.
\newblock In {\em Proceedings of AAAI}, pages 3776--3783.

\bibitem[\protect\citename{Shang \bgroup et al.\egroup }2015]{shang:2015}
Lifeng Shang, Zhengdong Lu, and Hang Li.
\newblock 2015.
\newblock Neural responding machine for short-text conversation.
\newblock In {\em Proceedings of ACL/IJCNLP}, pages 1577--1586.

\bibitem[\protect\citename{Traum}2003]{traum:2003}
David Traum.
\newblock 2003.
\newblock Issues in multiparty dialogues.
\newblock {\em Advances in Agent communication}, pages 201--211.

\bibitem[\protect\citename{Uthus and Aha}2013]{uthus:2013}
David~C Uthus and David~W Aha.
\newblock 2013.
\newblock Multiparticipant chat analysis: A survey.
\newblock {\em Artificial Intelligence}, pages 106--121.

\bibitem[\protect\citename{Vinyals and Le}2015]{vinyals:2015}
Oriol Vinyals and V.~Quoc Le.
\newblock 2015.
\newblock A neural conversational model.
\newblock \textit{arXiv preprint arXiv: 1506.05869}.

\bibitem[\protect\citename{Wang \bgroup et al.\egroup }2013]{wang:2013}
Hao Wang, Zhengdong Lu, Hang Li, and Enhong Chen.
\newblock 2013.
\newblock A dataset for research on short-text conversations.
\newblock In {\em Proceedings of EMNLP}, pages 935--945.

\bibitem[\protect\citename{Wang \bgroup et al.\egroup }2015]{wang:2015}
Mingxuan Wang, Zhengdong Lu, Hang Li, and Qun Liu.
\newblock 2015.
\newblock Syntax-based deep matching of short texts.
\newblock In {\em Proceedings of IJCAI}, pages 1354--1361.

\end{thebibliography}

\appendix

\section{Hyper-Parameters}
\label{sec:hp}

\begin{table}[H]
  \begin{center}
    {\small
  \begin{tabular}{lll} \toprule\hline
     Hyper-parameter  &  Values  \\ \hline
     Embedding size        &  512  \\
     GRU state size        &  256  \\
     WGAN $\lambda$        &  0.50    \\
     WGAN iterations       &  5  \\
     Critic hidden size    &  512 \\
     Critic activation function &  ReLU \\
     Batch size            &  32 \\
     Max epoch             &  30 \\
     Adam alpha            & \{0.001, 0.0005, 0.0001\} \\
     L2 weight decay       & \{0.001, 0.0005, 0.0001\}
                                    \\ \toprule
  \end{tabular}}
  \end{center}
  \caption{\label{tab:hyperparams} Hyper-parameters for our experiments.}
\end{table}


\section{Statics of Dataset}
Tab. \ref{tab:corpus} shows the details of our dataset.
``Docs'' is documents, ``Utters'' is utterances, ``W. / U.'' is the number of words per utterance, ``A. / D.'' is the number of agents per document.
\begin{table*}[h]
  \begin{center}
    {\small
  \begin{tabular}{l|ccccc} \toprule
    & \multicolumn{5}{c}{Dataset {\scriptsize (Train / Dev / Test)}} \\
                            & English ({\tt En})  &  Italian ({\tt It})  & Croatian ({\tt Hr})  & German ({\tt De}) & Portuguese ({\tt Pt}) \\ \midrule
    \multirow{2}{*}{No. of Docs}    & 7355     &  357     & 254      & 248       & 211  \\
                                    & {\scriptsize (6,606/ 367/ 382)}
                                    & {\scriptsize (306/ 17/ 34)}
                                    & {\scriptsize (216/ 12/ 26)}
                                    & {\scriptsize (216/ 12/ 20)}
                                    & {\scriptsize (180/ 10/ 21)}
                                    \\
    \multirow{2}{*}{No. of Utters}  & 2.4 M    &  165 k  &   80 k  &  38 k  & 52 k  \\
                                    & {\scriptsize (2.1 M / 13.2 k / 15.1 k)}
                                    & {\scriptsize (144  k / 7 k / 14 k)}
                                    & {\scriptsize ( 71  k / 3.4 k / 6.9 k)}
                                    & {\scriptsize ( 33  k / 1.9 / 2.9 k)}
                                    & {\scriptsize ( 44  k / 2.1 / 6.1 k )}
                                    \\
    \multirow{2}{*}{No. of Words}   & 27.0 M  & 1.1 M  &  630 k  &  335 k & 285 k  \\
                                    & {\scriptsize (23.8 M/ 1.5 M/ 1.7 M)}
                                    & {\scriptsize ( 1.0 M/  54 k/ 100 k)}
                                    & {\scriptsize ( 553 k/  25 k/  52 k)}
                                    & {\scriptsize ( 294 k/  16 k/  24 k)}
                                    & {\scriptsize ( 243 k/  11 k/  30 k)}
                                    \\
    \multirow{2}{*}{No. of Samples}
                                    & -   & -   & -   & -   & -      \\
                                    & {\scriptsize 665.6 k/ 45.1 k/ 51.9 k}
                                    & {\scriptsize 38511 / 2561 / 3873}
                                    & {\scriptsize 11387 /  512 / 1145}
                                    & {\scriptsize  5500 /  354 /  569}
                                    & {\scriptsize  5951 /  285 /  975}
                                    \\
    \multirow{2}{*}{Avg. W. / U.}   & 
                                       11.1 & 7.2 & 7.5 & 8.5 & 5.2  \\
                                    & {\scriptsize (11.1/ 11.2/ 11.3)}
                                    & {\scriptsize ( 6.9/  7.7/  7.1)}
                                    & {\scriptsize ( 7.7/  7.3/  7.5)}
                                    & {\scriptsize ( 8.9/  8.4/  8.2)}
                                    & {\scriptsize ( 5.5/  5.2/  4.9)}
                                    \\
    \multirow{2}{*}{Avg. A. / D.}   & 
                                       26.8  & 25.6  &  12.9 & 16.4 &  19.0 \\
                                    & {\scriptsize (26.3/ 30.68/ 32.1)}
                                    & {\scriptsize (24.9/ 26.2/ 25.6)}
                                    & {\scriptsize (12.7/ 13.5/ 12.7)}
                                    & {\scriptsize (17.4/ 15.9/ 15.8)}
                                    & {\scriptsize (19.7/ 18.6/ 18.8)}
                                    \\ \bottomrule
  \end{tabular}}
  \end{center}
  \caption{\label{tab:corpus} Statistics of the multilingual dataset. }
\end{table*}

\section{Results for each language}

Tab. \ref{tab:result_single} shows the detailed results of single-language adaptation.
Tab. \ref{tab:result_two} shows the detailed results of two-language adaptation.
Tab. \ref{tab:result_five} shows the detailed results of five-language adaptation.

\begin{table*}[h]
  {\small
  \begin{center}
  \begin{tabular}{c|c|l|ccc|ccc} \toprule
    & & & \multicolumn{3}{c|}{$|\mathcal{R}|=2$} & \multicolumn{3}{c}{$|\mathcal{R}|=10$} \\
    Target & \#Train &Method & {\tt ADR-RES} & {\tt ADR} & {\tt RES} & {\tt ADR-RES} & {\tt ADR} & {\tt RES} \\ \midrule
    \multirow{7}{*}{\tt It}
    & \multirow{7}{*}{38,511}
       & \sc Chance               & 2.99      &  5.97     &  50.00    &  0.60     &  5.97     &  10.00   \\
    &  & \sc TF-IDF               & 43.89     &  67.49	  &  64.58	  &  16.63	  &  67.49	  &  23.42 \\
    &  & \sc TrgOnly              & 63.28     &  79.86	  &  78.36    &  32.87    &  80.92	  &  38.73 \\
    &  & \sc EnOnly               & 44.54     &  72.37    &  60.11    &  9.91     &  66.74    &  16.29 \\
    &  & \sc FineTune             & 64.81     &  80.79    &  79.14    &  34.57    &  81.44    &  41.26 \\
    &  & \sc Joint                & 63.44     &  79.81    &  78.36    &  34.37    &  80.30    &  40.82 \\
    &  & \sc Wgan                 & 65.17     &  80.56    &  79.94    &  35.71    &  80.76    &  42.14 \\\midrule
    \multirow{7}{*}{\tt Hr}
    & \multirow{7}{*}{11,387}
       & \sc Chance               & 5.39      &  10.78    &  50.00    &  1.08     &  10.78    &  10.00 \\
    &  & \sc TF-IDF               & 35.63     &  58.78    &  61.05    &  10.22    &  58.78    &  17.29 \\
    &  & \sc TrgOnly              & 40.52     &  63.06    &  64.10    &  14.32    &  63.23    & 22.97 \\
    &  & \sc EnOnly               & 34.06     &  60.87    &  54.24    &  7.95	    &  60.52	  & 12.31  \\
    &  & \sc FineTune             & 40.00     &  62.62    &  63.23    &  14.67    &  64.72    & 22.79      \\
    &  & \sc Joint                & 44.37     &  62.97    &  69.26    &  15.98    &  62.97    & 24.54 \\
    &  & \sc Wgan                 & 45.07     &  62.62    &  70.48    &  16.59    &  63.76    & 25.68 \\\midrule
    \multirow{7}{*}{\tt De}
    & \multirow{7}{*}{5,500}
       & \sc Chance               &  4.09     &  8.17     &  50.00    &  0.82     &  8.17     &  10.00 \\
    &  & \sc TF-IDF               & 36.38     &  64.67    &  55.36    &  10.19    &  64.67    &  14.41\\
    &  & \sc TrgOnly              & 43.94     &  67.49    &  64.15    & 16.52     &  66.96    &  22.50 \\
    &  & \sc EnOnly               & 36.56     &  63.27    &  59.75    & 5.98      &  57.12    &  12.13 \\
    &  & \sc FineTune             & 50.44     &  68.89    &  72.06    & 20.39     & 68.19     &  26.71 \\
    &  & \sc Joint                & 50.79     &  70.30    &  70.47    & 21.79     & 69.24     &  27.94  \\
    &  & \sc Wgan                 & 52.90     &  71.53    &  72.23    & 22.50     & 68.89     & 28.30 \\\midrule
    \multirow{7}{*}{\tt Pt}
    &\multirow{7}{*}{5,951}
       & \sc Chance               & 3.42      &  6.84     &  50.00    &  0.68     &  6.84     &  10.00 \\
    &  & \sc TF-IDF               & 42.15     &  68.92    &  61.44    &  13.13    &  68.92    &  18.87  \\
    &  & \sc TrgOnly              & 41.64     &  66.67    &  62.77    & 13.95     &  67.79	  &  20.31  \\
    &  & \sc EnOnly               & 37.13     &  66.36    &  56.51    & 10.15     &  65.13    &  14.26 \\
    &  & \sc FineTune             & 43.08     &  66.05    &  64.92    & 14.97     &  66.97    &  21.85 \\
    &  & \sc Joint                & 47.59     &  68.10    &  69.44    & 17.13     &  68.92    &  24.21 \\
    &  & \sc Wgan                 & 49.54     &  69.23    &  70.36    & 18.56     &  67.38    &  25.44  \\\midrule
    \multirow{7}{*}{Avg.}
    &  \multirow{7}{*}{-}

       & \sc Chance               & \,\,\,3.97	&  \,\,\,7.94	& 50.00	& \,\,\,0.80	& \,\,\,7.94	& 10.00 \\
    &  & \sc TF-IDF               & 39.51	& 64.97	& 60.61	& 12.54	& 64.97	& 18.50 \\
    &  & \sc TrgOnly              & 47.35	& 69.27	& 67.35	& 19.42	& 69.73	& 26.13 \\
    &  & \sc EnOnly               & 38.07	& 65.72	& 57.65	& \,\,\,8.50	& 62.38	& 13.75 \\
    &  & \sc FineTune             & 49.58	& 69.59	& 69.84	& 21.15	& 70.33	& 28.15 \\
    &  & \sc Joint                & 51.55	& 70.30	& 71.88	& 22.32	& \bf 70.36	& 29.38 \\
    &  & \sc Wgan                 & \bf 53.17	& \bf 70.99	& \bf 73.25	& \bf 23.34	& 70.20	& \bf 30.39 \\
     \bottomrule
  \end{tabular}
  \end{center}}
  \vspace{-0.1cm}
  \caption{\label{tab:result_single} Results for single-language adaptation. Each number represents accuracy on addressee-response selection ({\tt ADR-RES}), addressee selection ({\tt ADR}) or response selection ({\tt RES}). \#Train is the number of training data.}
\end{table*}

\begin{table*}[t]
  {\small
  \begin{center}
  \begin{tabular}{c|l|ccc|ccc} \toprule
    & & \multicolumn{3}{c|}{ $|\mathcal{R}|=2$} & \multicolumn{3}{c}{ $|\mathcal{R}|=10$} \\
    Target & Method & {\tt ADR-RES} & {\tt ADR} & {\tt RES} & {\tt ADR-RES} & {\tt ADR} & {\tt RES} \\ \midrule
    \multirow{6}{*}{{\tt En}, {\tt It}}
    & \sc Chance               & \,\,\,1.80 {\scriptsize (\,\,\,0.62, \,\,\,2.95)}  &   \,\,\,3.61  & 50.00 &  \,\,\,0.36 {\scriptsize ( \,\,\,0.12,  \,\,\,0.60)}     &   3.61  & 10.00\\
    & \sc TF-IDF               & 40.51 {\scriptsize (37.13, 43.89)}  &  61.56  & 66.24 & 16.04 {\scriptsize (15.44, 16.63)}     &  61.56  & 26.31 \\
    & \sc EnOnly               & 50.01 {\scriptsize (55.47, 44.54)}  &  70.95  & 69.13 & 20.59 {\scriptsize (31.27, \,\,\,9.91)}     &  68.05	&  29.11 \\
    & \sc FineTune             & 59.13 {\scriptsize (53.44, 64.81)}  &  74.71  & 77.78 & 31.07 {\scriptsize (27.56, 34.57)}     &  74.62  &  39.34 \\
    & \sc Joint                & 59.56 {\scriptsize (55.67, 63.44)}  &  74.63  & 78.50 & 33.04 {\scriptsize (31.71, 34.37)}     &  74.77	&  41.72 \\
    & \sc Wgan                 & 60.20 {\scriptsize (55.23, 65.17)}  &  74.94  & 79.10 & 33.38 {\scriptsize (31.04, 35.71)}     &  75.09  &  41.87 \\\midrule
    \multirow{6}{*}{{\tt En}, {\tt Hr}}

    & \sc Chance               & \,\,\,2.35 {\scriptsize (\,\,\,0.62, \,\,\,5.39)}  &   4.71 & 50.00 & \,\,\,0.47 {\scriptsize (\,\,\,0.12, \,\,\,1.08)}  &  \,\,\,4.71 & 10.00 \\
    & \sc TF-IDF               & 36.76 {\scriptsize (37.13, 35.63)}  &  60.15 & 61.63 & 12.82 {\scriptsize (15.44, 10.22)}  &  60.15 & 21.80 \\
    & \sc EnOnly               & 44.77 {\scriptsize (55.47, 34.06)}  &  65.20 & 66.20 & 19.61 {\scriptsize (31.27, \,\,\,7.95)}  &  64.94 & 27.12 \\
    & \sc FineTune             & 46.03 {\scriptsize (52.06, 40.00)}  &  65.29 & 69.15 & 20.96 {\scriptsize (27.24, 14.67)}  &  66.01 & 30.28 \\
    & \sc Joint                & 49.49 {\scriptsize (55.66, 43.32)}  &  65.98 & 73.14 & 22.95 {\scriptsize (31.74, 14.15)}  &  66.15 & 32.23 \\
    & \sc Wgan                 & 49.80 {\scriptsize (54.53, 45.07)}  &  65.70 & 73.96 & 23.61 {\scriptsize (30.62, 16.59)}  &  66.38 & 33.52 \\\midrule
    \multirow{6}{*}{{\tt En}, \tt {De}}
    & \sc Chance               & \,\,\,3.01 {\scriptsize (\,\,\,0.62, \,\,\,4.08)}  & \,\,\,6.01 & 50.00  & \,\,\,0.60 {\scriptsize (\,\,\,0.12, \,\,\,0.82)}  &  \,\,\,6.01 & 10.00 \\
    & \sc TF-IDF               & 36.38 {\scriptsize (37.13, 36.38)}  &  57.20 & 64.47 & 12.83 {\scriptsize (15.44, 10.19)}  &  57.20 & 23.24 \\
    & \sc EnOnly               & 46.02 {\scriptsize (55.47, 36.56)}  &  66.40 & 68.95 & 18.63 {\scriptsize (31.27, \,\,\,5.98)}  &  63.24 & 27.03 \\
    & \sc FineTune             & 52.22 {\scriptsize (53.99, 50.44)}  &  68.79 & 74.52 & 24.66 {\scriptsize (28.93, 20.39)}  &  68.20 & 33.09 \\
    & \sc Joint                & 53.04 {\scriptsize (55.28, 50.79)}  &  69.73 & 74.30 & 26.09 {\scriptsize (31.97, 20.21)}  &  68.98 & 35.14 \\
    & \sc Wgan                 & 54.05 {\scriptsize (55.20, 52.90)}  &  70.35 & 75.19 & 27.05 {\scriptsize (31.42, 22.67)}  &  69.43 & 35.31 \\\midrule
    \multirow{6}{*}{{\tt En}, {\tt Pt}}
    & \sc Chance               & \,\,\,2.02 {\scriptsize (\,\,\,0.62, \,\,\,3.42)}  & \,\,\,4.04  & 50.00 & \,\,\,0.40 {\scriptsize (\,\,\,0.12, \,\,\,0.68)}  &  \,\,\,4.04 & 10.00 \\
    & \sc TF-IDF               & 39.64 {\scriptsize (37.13, 42.15)}  &  62.27 & 64.67 & 14.29 {\scriptsize (15.44, 13.13)}  &  62.27 & 24.03 \\
    & \sc EnOnly               & 46.30 {\scriptsize (55.47, 37.13)}  &  67.94 & 67.33 & 20.71 {\scriptsize (31.27, 10.15)}  &  67.24 & 28.09 \\
    & \sc FineTune             & 46.53 {\scriptsize (49.98, 43.08)}  &  66.39 & 68.97 & 20.52 {\scriptsize (26.06, 14.97)}  &  66.74 & 29.13 \\
    & \sc Joint                & 51.39 {\scriptsize (55.18, 47.59)}  &  68.66 & 73.81 & 24.32 {\scriptsize (31.51, 17.13)}  &  69.11 & 33.28 \\
    & \sc Wgan                 & 52.49 {\scriptsize (55.44, 49.54)}  &  69.31 & 74.29 & 24.88 {\scriptsize (31.19, 18.56)}  &  68.16 & 33.75 \\\midrule
    \multirow{6}{*}{Avg.}
     & \sc Chance              & \,\,\,2.30	 &  \,\,\,4.59 &  50.00 & \,\,\,0.46  &  	\,\,\,4.59 & 	10.00\\
     & \sc TF-IDF              & 38.32 & 60.29 &  64.25 & 13.99 &  60.29 &  23.84  \\
     & \sc EnOnly              & 46.77 & 67.62 & 67.90 & 19.88 & 65.86 & 27.83 \\
     & \sc FineTune            & 50.98 & 68.79 & 72.60 & 24.30 & 68.89 & 32.96\\
     & \sc Joint               & 53.37 & 69.75 & 74.94 & 26.60 & 69.75 & 35.59\\
     & \sc Wgan                & \bf 54.14 & \bf 70.07 & \bf 75.63 & \bf 27.23 & \bf 69.76 & \bf 36.11 \\ \midrule

     \bottomrule
  \end{tabular}
  \end{center}}
  \vspace{-0.1cm}
  \caption{\label{tab:result_two} Results for two-language adaptation. Each number is macro average of the accuracies over all the languages. Each of the parenthesized numbers in the {\tt ADR-RES} column is {\tt ADR-RES} accuracy for each language.}
\end{table*}

\begin{table*}[t]
  {\small
  \begin{center}
  \begin{tabular}{c|l|ccc|ccc} \toprule
    & & \multicolumn{3}{c|}{ $|\mathcal{R}|=2$} & \multicolumn{3}{c}{ $|\mathcal{R}|=10$} \\
    Target & Method & {\tt ADR-RES} & {\tt ADR} & {\tt RES} & {\tt ADR-RES} & {\tt ADR} & {\tt RES} \\ \midrule
    \multirow{6}{*}{{\tt En}, {\tt It}}

    & \multirow{2}{*}{\sc TF-IDF}
                                &  39.04 & 63.10 & 62.06 & 13.12 & 63.10 & 20.64 \\
    &                           & {\scriptsize (37.13, 43.89, 35.63, 36.38, 42.15)} &       &       & {\scriptsize (15.44, 16.63, 10.22, 10.19, 13.13)}  &        &       \\

    & \multirow{2}{*}{\sc EnOnly}
                                &  41.55 & 66.48 & 61.75 & 13.05 & 63.77 & 19.38 \\
    &                           & {\scriptsize (55.47, 44.54, 34.06, 36.56, 37.13)} &       &       & {\scriptsize (31.27, \,\,\,9.91, \,\,\,7.95, \,\,\,5.98, 10.15)}  &        &       \\
      \multirow{2}{*}{{\tt Hr}, {\tt De}, {\tt Pt}}
    & \multirow{2}{*}{\sc Joint} & 50.69 & 69.00 & 72.18 & 22.80 & 69.18 & 31.11 \\
    &                            & {\scriptsize (54.49, 61.71, 43.41, 47.80, 46.05)} &       &       & {\scriptsize (31.26, 30.29, 14.59, 20.21, 17.64)}  &        &       \\
    & \multirow{2}{*}{\sc Wgan} & \bf 52.11 & \bf 69.74 & \bf 73.34 & \bf 23.39 & \bf 69.35 & \bf 31.88 \\
    &                           & {\scriptsize (53.88, 63.18, 44.19, 52.02, 47.28)} &       &       & {\scriptsize (30.6, 31.29, 14.67, 22.32, 18.05)}  &        &       \\
     \bottomrule
  \end{tabular}
  \end{center}}
  \vspace{-0.1cm}
  \caption{\label{tab:result_five} Results for five-language adaptation. Each number is macro average of the accuracies over all the languages. Each of the parenthesized numbers in the {\tt ADR-RES} column is {\tt ADR-RES} accuracy for each language.}
\end{table*}

\end{document}